\pdfoutput=1

\documentclass[11pt]{article}

\usepackage[preprint]{acl}

\usepackage{times}
\usepackage{latexsym}
\usepackage{tabularx}
\usepackage[T1]{fontenc}

\usepackage[utf8]{inputenc}

\usepackage{microtype}

\usepackage{inconsolata}

\usepackage{graphicx}

\usepackage[utf8]{inputenc}
\usepackage{CJKutf8}
\usepackage{booktabs}
\usepackage{tcolorbox}
\usepackage{multirow}
\usepackage{arydshln}
\usepackage{bm}

\title{Unveiling Cultural Blind Spots: Analyzing the Limitations of mLLMs\\ in Procedural Text Comprehension}

\author{
 \textbf{Amir Hossein Yari\textsuperscript{1}} \qquad
 \textbf{Fajri Koto\textsuperscript{2}}
\\
\\
 \textsuperscript{1}Sharif University of Technology\\
 \textsuperscript{2}Department of Natural Language Processing, MBZUAI
\\
\small \texttt{ahyari2002@gmail.com}, \texttt{fajri.koto@mbzuai.ac.ae} \normalsize
}

\begin{document}
\maketitle
\begin{abstract}
Despite the impressive performance of multilingual large language models (mLLMs) in various natural language processing tasks, their ability to understand procedural texts, particularly those with culture-specific content, remains largely unexplored. Texts describing cultural procedures, including rituals, traditional craftsmanship, and social etiquette, require an inherent understanding of cultural context, presenting a significant challenge for mLLMs. In this work, we introduce \textbf{\texttt{CAPTex}}, a benchmark designed to evaluate mLLMs' ability to process and reason about culturally diverse procedural texts across multiple languages using various methodologies to assess their performance. Our findings indicate that (1) mLLMs face difficulties with culturally contextualized procedural texts, showing notable performance declines in low-resource languages, (2) model performance fluctuates across cultural domains, with some areas presenting greater difficulties, and (3) language models exhibit better performance on multiple-choice tasks within conversational frameworks compared to direct questioning. These results underscore the current limitations of mLLMs in handling culturally nuanced procedural texts and highlight the need for culturally aware benchmarks like \textbf{\texttt{CAPTex}} to enhance their adaptability and comprehension across diverse linguistic and cultural landscapes.\footnote{This dataset will be publicly released under a Creative Commons license at \url{https://huggingface.co/datasets/AmirHossein2002/CAPTex}}
\end{abstract}
\section{Introduction}

Procedural texts encompass a genre of writing that provides systematic instructions or guidance to navigate a sequence of actions or steps, aiming to achieve a specific outcome. These texts are common in various fields, including technical documentation, user manuals, and cookbooks. The core characteristic of procedural texts is their sequential and organized structure, with each instruction building on the previous one to ensure readers can successfully reach the intended outcome. Unlike other writing styles such as narrative or descriptive, procedural texts emphasize clarity, accuracy, and a straightforward progression of actions to enable effective task completion.

Large Language Models (LLMs) have demonstrated exceptional capabilities across various natural language processing (NLP) tasks, such as text summarization \citep{jin2024comprehensive}, multi-modal machine translation \citep{shen2024surveymultimodalmachinetranslation}, solving complex tasks modeled as state machines \citep{wu2024stateflow}, and code generation and understanding \citep{e25060888}. Unlike traditional models that rely on task-specific training, LLMs can be adapted to a wide range of applications through effective prompting strategies, making them suitable for diverse and dynamic contexts \citep{ouyang2022traininglanguagemodelsfollow, dai2023can}.

One particularly significant application area for LLMs is their ability to accurately interpret procedural texts. This capability is becoming increasingly vital as these models are employed in tasks like generating automated instructions and facilitating human-computer interactions \citep{Kosch_2024}. In such scenarios, the demand for clear, contextually appropriate, and executable steps is critical. However, inaccuracies or ambiguities in interpreting procedural instructions can result in miscommunication, errors, and inefficiencies. These issues are particularly concerning in real-world domains such as healthcare, education, and technical support, where precision and clarity are paramount. Consequently, ensuring that LLMs can reliably process procedural texts is essential for their effective and responsible integration into various systems.

Culture plays a fundamental role in both the creation and comprehension of procedures. A step-by-step instruction in one culture may rely on shared knowledge or tacit understanding that is absent in others \citep{ca4e4e63-bd4b-338b-987d-87397047284e}. For instance, instructions on performing a ritual practice may have meanings, symbols, and steps deeply embedded in the traditions of a particular culture, as demonstrated in Table \ref{tab:funeral}. If language models are unable to recognize and navigate these cultural dimensions, they risk misrepresenting the intent or structure of the procedure, leading to errors or misinterpretations.

\begin{table}
  \centering
  \resizebox{\linewidth}{!}{
  \begin{tabular}{wc{0.4cm} p{3cm} p{3.2cm}}
    \toprule
    \textbf{Step} & \textbf{Iran} & \textbf{Indonesia} \\
    \midrule
    \centering 1     & \small The body is taken to the cemetery for burial. & \small The traditional Toraja house is prepared in silence, and tools are gathered.           \\
    \midrule
    \centering 2     & \small The deceased's body is washed according to Islamic rituals. & \small The body is wrapped, and the coffin is decorated in a ritual performance.          \\
    \midrule
    \centering 3     & \small The body is wrapped in a plain white shroud. & \small A cultural parade is held, transporting the body from the house to the burial site.           \\
    \midrule
    \centering 4      & \small A special prayer is performed for the deceased. & \small A traditional Toraja dance is performed as part of the ceremony.            \\
    \midrule
    \centering 5      & \small The body is buried facing Mecca, with a layer of dirt and stones over the grave.   & \small Animal sacrifices, typically buffalo and pigs, are offered as part of the final rites.       \\
    \bottomrule
  \end{tabular}
  }
  \caption{Comparison of funeral practices in Iran and Indonesia. For Indonesia, it's a tradition from North Sumatra.}
  \label{tab:funeral}
\end{table}
\normalsize

Studies have shown that current large language models (LLMs) tend to exhibit biases favoring Western perspectives, mirroring the cultural norms and values of Western, educated, industrialized, rich, and democratic (WEIRD) societies \citep{durmus2024measuringrepresentationsubjectiveglobal, Naous2023HavingBA}, while often inadequately representing other cultural contexts \citep{ prabhakaran2022culturalincongruenciesartificialintelligence}. These biases primarily stem from the nature of training data \citep{arora-etal-2022-exposure, 10.1145/3531146.3533229, nadeem-etal-2021-stereoset} and design decisions, including model architecture, tokenization approaches, evaluation methods, and instruction-tuning techniques. The significance of LLMs accurately comprehending these texts extends beyond the technical understanding of instructions; it also involves ensuring fairness, accessibility, and cultural sensitivity.  Evaluating how well LLMs can decode procedural texts, particularly when cultural context plays a pivotal role, is critical for advancing their capability to serve a diverse range of users across different linguistic and cultural backgrounds.

Cultural procedural texts, which are deeply intertwined with societal norms, values, and traditions, pose unique challenges for LLMs. This complexity raises several critical questions about the ability of LLMs to effectively navigate and reason within culturally specific contexts:  (1) How do LLMs perform in understanding procedural texts in low-resource languages compared to high-resource languages? (2) How effectively can LLMs recognize, interpret, and preserve the cultural nuances embedded in procedural texts? (3) Do LLMs demonstrate consistent performance across different cultural domains, such as food preparation, religious rituals, and celebration setups? (4) Are there noticeable strengths or weaknesses in LLMs’ understanding depending on the cultural context of the procedural text?

To address these questions, we introduce \textbf{\texttt{CAPTex}} (\textbf{C}ulturally-\textbf{A}ware \textbf{P}rocedural \textbf{Tex}ts), an innovative dataset crafted to evaluate multilingual LLMs' (mLLMs) ability to reason culturally through the lens of procedural text understanding across diverse tasks, including reordering tasks, multiple-choice questions, and conversational frameworks, each of which is elaborated upon in detail in Section \ref{sec:tasks_formulation}. \textbf{\texttt{CAPTex}} is carefully developed with contributions from native speakers representing seven culturally distinct regions—China, India, Indonesia, Iran, Japan, Nigeria, and Pakistan—ensuring authentic and nuanced cultural representation.

\section{Related Work}

Procedural text analysis has been a focal point of research, addressing a wide array of tasks within this domain. For example, \citet{cao2023culturaladaptationrecipes} tackles the cultural adaptation of recipes between Chinese and English-speaking cuisines. Their work aims to automate the translation and cultural adaptation of recipes, ensuring that cultural nuances—including ingredients, cooking techniques, and unit conversions—are appropriately represented. In contrast, our work extends beyond the food domain, encompassing multiple cultural contexts across seven countries, thereby offering a broader perspective on cross-cultural procedural knowledge. 

Several studies have also focused on advancing entity tracking methodologies. NCET \citep{gupta-durrett-2019-tracking} introduces a mechanism for continuous-space entity tracking, employing a conditional random field (CRF) to ensure sequential consistency in predictions. Similarly, \citet{huang-etal-2021-reasoning} utilizes a graph neural network to model semantic relationships among entities, actions, and locations, enhancing the understanding of procedural text.

Incorporating temporal aspects into procedural comprehension, \citet{rajaby-faghihi-kordjamshidi-2021-time} propose the Time-Stamped Language Model (TSLM), which augments pre-trained language models with timestamp embeddings. This approach has significantly improved performance on datasets such as Propara \cite{dalvi-etal-2018-tracking} and NPN-Cooking. Additionally, \citep{tang-etal-2020-understanding-procedural} introduces the Interactive Entity Network (IEN), a recurrent network with memory designed to capture diverse entity interactions for state tracking. Meanwhile, \citet{amini2020proceduralreadingcomprehensionattributeaware} develops an algorithm for procedural reading comprehension, translating texts into a formalism that represents processes as sequences of transitions over entity attributes.

Efforts to integrate multimodal data have also advanced procedural text analysis. For instance, \citep{10.1145/3397271.3401247} introduces a transformer-based model that combines textual and visual information for processing multimodal recipe datasets effectively. Building on this, \citep{wu-etal-2022-understanding} conducts benchmarking on reasoning and sequencing unordered multimodal instructions, highlighting that state-of-the-art models still fall short of human-level performance. While their work primarily focuses on step reordering, our evaluation framework is more comprehensive, introducing three additional tasks to assess LLMs’ capabilities. Furthermore, rather than being restricted to English, our research incorporates the native languages of the targeted countries, ensuring that the procedures analyzed are culturally unique rather than globally common.

Despite these advancements, a holistic benchmark for procedural text comprehension remains elusive. Our work sets a new standard by extending beyond food-related tasks to encompass multiple domains, incorporating a diverse range of languages beyond English, and evaluating the capabilities and limitations of mLLMs through a multifaceted assessment framework. In the following section, we will elaborate on these methodologies in detail, highlighting how our benchmark surpasses prior efforts.

\section{CAPTex}

To address our research objectives, we introduce \textbf{\texttt{CAPTex}}, a dataset that incorporates cultural procedural texts across English and seven additional languages from diverse linguistic and geographical backgrounds. \textbf{\texttt{CAPTex}} is built upon three foundational components: (1) a curated collection of procedures spanning ten unique categories, (2) a series of thoughtfully crafted multiple-choice questions designed to evaluate the comprehension of each procedure, and (3) a rich corpus of conversational exchanges offering clarifications on the corresponding procedures.

\begin{figure*}[t]
  \includegraphics[width=2.1\columnwidth]{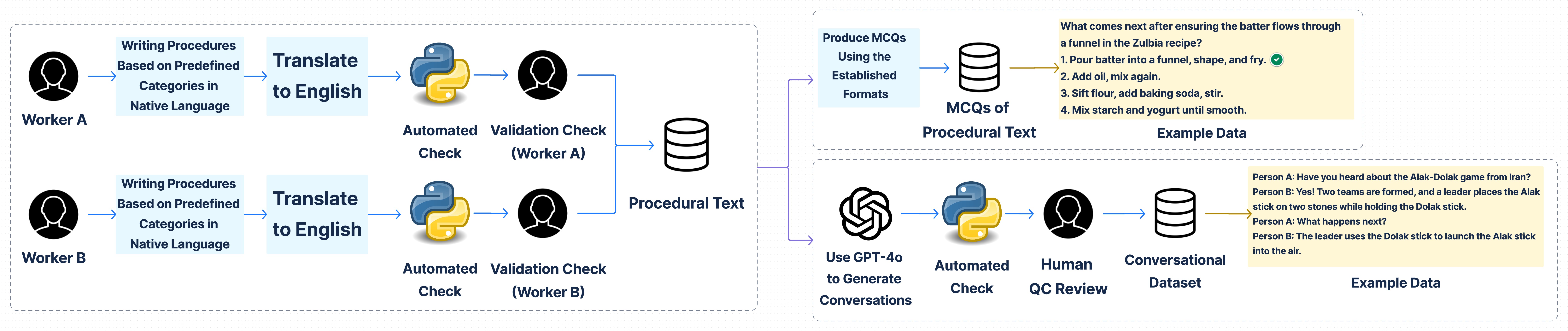}
  \caption{End-to-End process of dataset creation}
  \label{fig:dataset_creation}
\end{figure*}

\subsection{Data Construction}

\textbf{Language Coverage} Contemporary large language models demonstrate impressive performance in languages with abundant training data; however, their capabilities diminish when applied to low-resource languages and intricate cultural contexts, thereby constraining their global applicability \citep{maaz2024palopolyglotlargemultimodal}. To promote linguistic diversity, we selected languages—Chinese (Mandarin), Japanese, Persian, Hindi, Indonesian, Urdu, and Hausa—representing a spectrum of resource availability, as measured by the criteria established by \citet{joshi-etal-2020-state}. Table~\ref{tab:resource_availability} presents a detailed summary of the resource availability for each language featured in \textbf{\texttt{CAPTex}}.
\begin{table}
  \centering
  \resizebox{0.8\linewidth}{!}{
  \begin{tabular}{ll}
    \toprule
    \textbf{Language} & \textbf{Class} \\
    \hline
    Chinese (Mandarin)     & 5 - The Winners           \\
    Japanese     & 5 - The Winners          \\
    Persian     & 4 - The Underdogs           \\
    Hindi      & 4 - The Underdogs            \\
    Indonesian      & 3 - The Rising Stars           \\
    Urdu      & 3 - The Rising Stars           \\
    Hausa     & 2 - The Hopefuls          \\
    \bottomrule
  \end{tabular}
  }
  \caption{Resource availability of languages in \textbf{\texttt{CAPTex}}}
  \label{tab:resource_availability}
\end{table}

\paragraph{Topic Taxonomies} The procedures in \textbf{\texttt{CAPTex}} span ten culturally significant domains: (1) food and cuisine, (2) celebrations and festivals, (3) social etiquette and hospitality, (4) craftsmanship and artisan skills, (5) traditional attire and dress, (6) agricultural and seasonal practices, (7) religious and spiritual practices, (8) life milestones and family rites, (9) sports, games, and competitions, and (10) environmental and nature-based practices. These categories were selected by adapting and expanding upon the taxonomy from \texttt{IndoCulture} \citep{koto-etal-2024-indoculture} to ensure comprehensive coverage of cultural traditions and everyday practices. Please refer to Appendix for more detailed topic taxonomies.

\paragraph{Writing Procedural Text} 
As shown in Figure~\ref{fig:dataset_creation}, for each language, we employed two native speakers from each target nation to manually write procedural texts in their native language along with their English translations. These workers have deep cultural ties, having spent their entire lives in their native lands, ensuring strong familiarity with local traditions. The only exceptions are two individuals who lived in their home countries for the first 25 years before moving abroad, but have been residing outside their homeland for fewer than five years.\footnote{Each worker is compensated fairly based on a five-day workload, with payments aligned with the minimum wage in their respective country.}

Given a specific topic or category, each worker manually wrote procedural texts in their native language and translated them into English. They were strictly prohibited from using AI-based text generation tools but were allowed to reference reliable literature to verify cultural details and improve the accuracy of their writing. In total, we produce 1,400 human-written procedural texts (100 texts per language × 7 native languages × 2, including English translations).

\paragraph{Quality Control} We ensure the high quality of \textbf{\texttt{CAPTex}} through two quality checks. First, we conduct an automated check to verify that each procedure consists of 5–10 steps and that the step count aligns between the native language and its English translation. Second, we perform a manual review by having workers cross-check procedural texts written by their peers. This evaluation follows a detailed checklist assessing conceptual accuracy, cultural relevance, logical progression, step order, grammatical correctness, and the accuracy and consistency of English translations. Any issues identified during the manual review are addressed by having the original worker revise their text accordingly. A common challenge arises from the lack of logical progression, which requires consolidating certain steps or eliminating those that do not impact the sequence. This restructuring makes the order of the remaining steps critical, as they become inherently non-interchangeable. For a more detailed description of the data collection process and quality control measures, refer to Appendix \ref{sec:appendix:data_collection}.

\subsection{Task Formulation}
\label{sec:tasks_formulation}

Using \textbf{\texttt{CAPTex}}, we developed four tasks, each designed with specific objectives. The following sections provide a detailed explanation of these tasks and their intended goals.

\paragraph{Task 1: Step Reordering} In this task, procedural steps are initially shuffled and labeled with sequential letters (such as A, B, C, D). The model is then tasked with reconstructing the correct sequence, outputting a comma-separated list of these letters without any additional explanation.

To gauge performance, we utilize three established metrics. The first, Spearman's rank correlation \citep{ca468a70-0be4-389a-b0b9-5dd1ff52b33f}, examines the monotonic relationship between the predicted and actual rankings. The second, Levenshtein distance \citep{Levenshtein1965BinaryCC}, measures the minimum number of edit operations needed to transform the predicted sequence into the correct one. Lastly, Kendall's Tau rank correlation \citep{665905b2-6123-3642-832e-05dbc1f48979} assesses the ordinal agreement between the predicted and true sequence by counting pairwise swaps. This assessment is performed in both English and a native language, ensuring a thorough evaluation of the model's ability to generalize across different linguistic environments. 

\paragraph{Task 2: Procedure-Based Multiple-Choice Questions (PB-MCQ)} 
We design a comprehensive multiple-choice question (MCQ) framework for each procedure to evaluate mLLMs' comprehension and reasoning abilities in identifying both subsequent and preceding steps. We created affirmative and negative versions for each question type, constructing them in the original language as well as in English. This approach ensured consistency across all question types (Subsequent Affirmative, Subsequent Negative, Antecedent Affirmative, Antecedent Negative) in both languages. To maintain linguistic parity, we construct all questions in both the original language and English. Each question consists of four answer choices, with one correct option. As a result, we generate eight MCQs per procedure, leading to a total of 5,600 questions (4 question types × 100 procedures × 14 languages, including native languages and their English translations).

In our question formulation, we ensured that for queries about upcoming steps, the correct answer was the next step, while three incorrect options were randomly chosen from earlier steps. Likewise, for questions regarding previous steps, the correct response was the preceding step, with three incorrect choices randomly selected from later steps. We confirmed that each question had one correct answer and three incorrect options. An illustrative example of MCQs is provided in Table \ref{tab:mcq_example}. For additional information about MCQs, please refer to appendix \ref{sec:appendix:mcq_design}

The language model is prompted to generate only the correct choice option (A, B, C, or D) as its output. The primary evaluation metric for assessing model performance in this task is accuracy, measured by the proportion of correctly selected answers. Beyond evaluating the cultural understanding and procedural reasoning of language models, this task also enables an analysis of how well mLLMs comprehend affirmative and negative questions. Additionally, it assesses the model’s ability to predict both the subsequent and the antecedent procedural steps, further refining our understanding of its reasoning and contextual awareness.

\paragraph{Task 3: Conversation-Based Multiple-Choice Questions (CB-MCQ)}  To evaluate the reasoning capabilities of large language models (LLMs) using procedural text, we created procedurally grounded conversations in English. We utilized GPT-4o along with a specially designed prompt (detailed in Appendix \ref{sec:appendix:conversation}) to generate a series of four-utterance dialogues, conditioned on \textbf{\texttt{CAPTex}}. These dialogues simulate a natural conversation between two individuals, referred to as Person A and Person B. In the conversation, Person A starts by asking about a particular procedure. Person B, acting as a knowledgeable respondent, introduces the procedure and describes the initial steps involved. Person A then asks a follow-up question to clarify the next step in the process, and Person B provides a detailed explanation in response. Table \ref{tab:conversation_example} presents an example of the generated conversations.

Based on the conversation structure, the third utterance presents a question regarding the next step of the process. However, the fourth utterance, which contains the explanation of that step, is intentionally omitted. The language model must then select the correct answer from four options corresponding to the missing explanation, using the same answer choices as PB-MCQ. Comparing PB-MCQ and CB-MCQ allows us to assess whether models perform better in structured question-answering or conversational reasoning. PB-MCQ evaluates direct procedural knowledge, while CB-MCQ tests inference within dialogue, providing insights into model adaptability across different task formats.

For quality control, we employ a native speaker to review each dialogue assessing conceptual accuracy, grammatical correctness, and the consistency of Person B's explanations with the actual steps outlined in the procedure. If any errors are identified, the worker manually corrects them to maintain accuracy and coherence. Similar to PB-MCQ, we use accuracy as the evaluation metric for CB-MCQ.

\paragraph{Task 4: The Conversation-Based Question Answering (CB-QA)} This task is similar to CB-MCQ but differs in that it requires the model to generate a response as if it were the conversational participant, rather than selecting from predefined choices. This task is chosen over CB-MCQ to evaluate the model’s ability to produce natural, contextually appropriate responses, reflecting a deeper understanding of procedural knowledge in dialogue. The performance of mLLMs is evaluated using three metrics: ROUGE-L score \citep{lin-2004-rouge}, BERTScore \citep{zhang2020bertscoreevaluatingtextgeneration}, and additional semantic similarity score \citep{corley-mihalcea-2005-measuring}, which quantifies meaning alignment between generated and reference responses.

\subsection{Data Statistics}
\label{sec:data_stat}
\begin{figure}[t]
  \includegraphics[width=0.9\columnwidth]{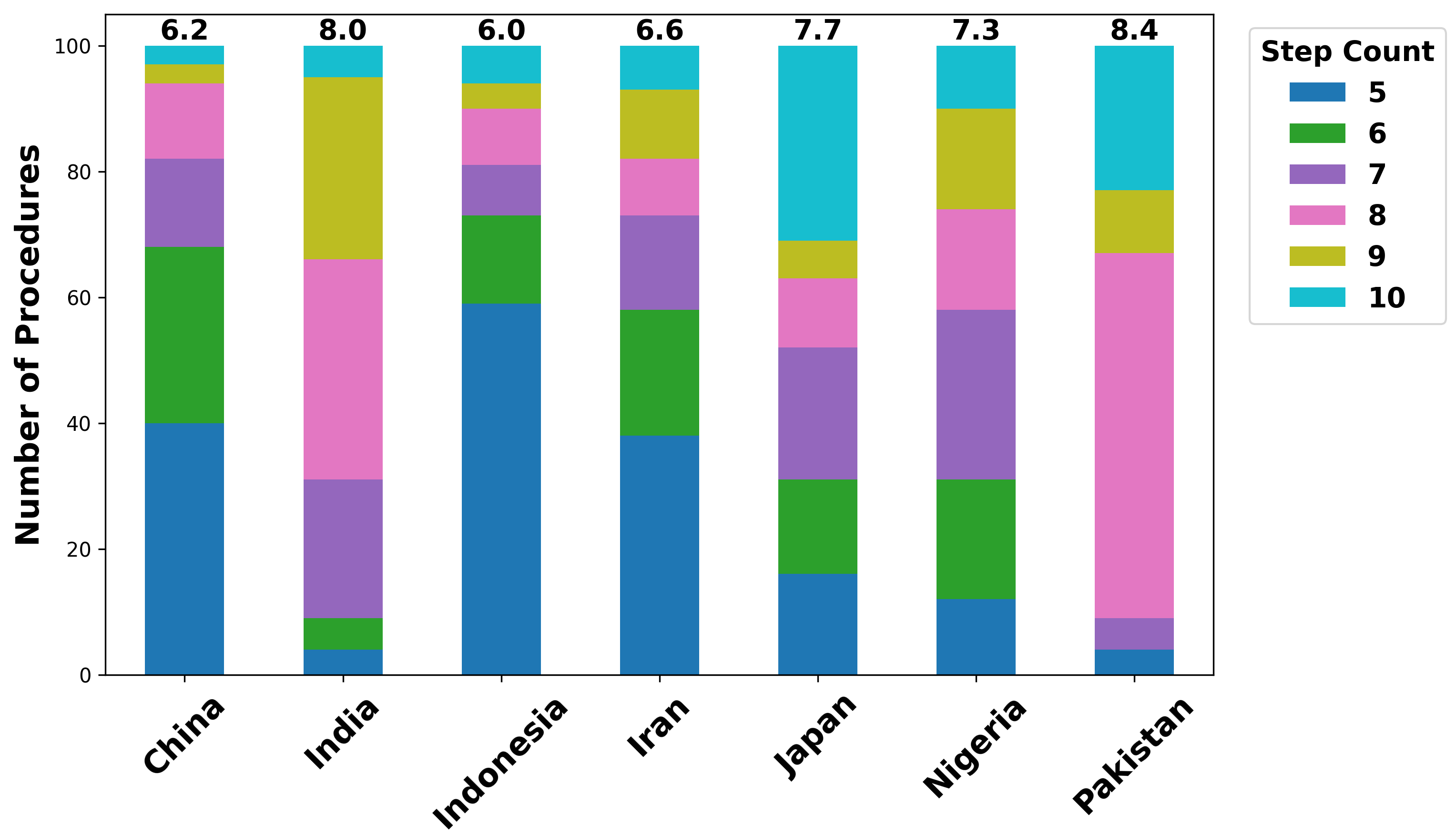}
  \caption{Procedures step counts by country}
  \label{fig:hist_country}
  \vspace{-0.5cm}
\end{figure}

For each country and category, \textbf{\texttt{CAPTex}} incorporates 10 procedures, totaling 100 distinct procedures per country. Figure \ref{fig:hist_country} represents the distribution of procedural steps across countries, with numbers above each bar indicating the average number of steps.

\begin{table*}
  \centering
    \resizebox{0.85\textwidth}{!}{
  \begin{tabular}{p{1.5cm} wc{2cm} wc{1.6cm} wc{2cm} wc{2cm} wc{0.01cm} wc{0.8cm} wc{1cm} wc{0.8cm} wc{1cm}}
    \toprule
     \multirow{2}{*}{\textbf{Country}} &  \multicolumn{1}{c}{\textbf{Procedures}} & \multicolumn{3}{c}{\textbf{MCQ} \footnotesize \textbf{(English/Native)}} & \multicolumn{5}{c}{\textbf{Conversations (Utterances)}} \\
    \cline{3-5} \cline{7-10}
    & \footnotesize \textbf{(English/Native)} & \small \textbf{Question} & \small \textbf{Correct Ans.} & \small \textbf{Incorrect Ans.} & & \small \textbf{First} & \small \textbf{Second} & \small \textbf{Third} & \small \textbf{Fourth} \\
    \hline
    China     & 157.8 / 138.8 & 38.0 / 39.8 & 24.5 / 21.2 & 26.0 / 22.8 & & 9.4 & 46.2 & 13.1 & 34.3  \\
    India     & 37.5 / 50.4   & 16.5 / 21.8 & 4.8 / 6.3   & 4.7 / 6.3 & & 9.3 & 37.2 & 12.7 & 23.6 \\
    Indonesia & 121.9 / 98.9   & 39.1 / 31.9 & 19.5 / 15.6 & 20.1 / 16.4 & & 14.8 & 41.1 & 12.4 & 30.1 \\
    Iran      & 144.5 / 166.4 & 34.3 / 40.3 & 21.6 / 24.7 & 22.4 / 25.8 & & 10.0 & 44.7 & 13.2 & 30.2 \\
    Japan     & 120.6 / 147.1 & 30.6 / 45.5 & 16.2 / 19.6 & 15.8 / 19.2 & & 12.3 & 43.8 & 13.1 & 29.5 \\
    Nigeria   & 162.3 / 173.3 & 35.8 / 38.1 & 21.1 / 22.8 & 21.8 / 23.5 & & 11.0 & 45.7 & 13.2 & 29.6 \\
    Pakistan  & 76.6 / 99.7 & 23.9 / 36.4 & 9.2 / 12.0  & 9.4 / 12.2 & & 12.0 & 42.0 & 12.9 & 27.0 \\
    \bottomrule
  \end{tabular}
  }
  \caption{Average word counts for \textbf{\texttt{CAPTex}} components (Procedures, MCQs, and Conversations) by country.}
  \label{tab:word_count_country}
\end{table*}

The methodology guarantees an equitable distribution of multiple-choice questions (MCQs) across various countries, categories, and formats. For each designated format\footnote{Subsequent Affirmative, Subsequent Negative, Antecedent Affirmative, and Antecedent Negative} a single question is crafted. This approach results in the creation of 2,800 unique questions in the original language and an additional 2,800 in English, culminating in a total of 5,600 MCQs. Tables \ref{tab:word_count_country} and \ref{tab:word_count_category} present the mean lexical density of \textbf{\texttt{CAPTex}}, analyzed across countries and categories, respectively.

The conversations dataset is thoroughly balanced, with one conversation constructed for each procedure. This approach yielded a total of 700 English conversations.

\section{Experiments}

\subsection{Setup}

We assessed 31 multilingual language models of different sizes, including DeepSeek\citep{deepseekai2025deepseekr1incentivizingreasoningcapability}, Gemma-2 \citep{gemmateam2024gemma2improvingopen}, Llama-3 \citep{grattafiori2024llama3herdmodels}, Mamba \citep{gu2024mambalineartimesequencemodeling}, Mistral \citep{jiang2023mistral7b}, Qwen2.5 \citep{qwen2025qwen25technicalreport}, BLOOMZ \citep{muennighoff-etal-2023-crosslingual}, Aya-Expanse \citep{dang2024ayaexpansecombiningresearch}, mT0 \citep{muennighoff-etal-2023-crosslingual}, GPT-4 \citep{openai2024gpt4technicalreport}, and O3-mini \citep{openai2025o3mini}.

We conducted zero-shot evaluations using prompt templates exclusively in English. Prior research has shown that prompting in different languages can lead to variations in responses to similar queries \citep{lin-etal-2022-shot, Shen2024TheLB}. Moreover, studies on multilingual LLMs have consistently found that these models tend to perform better when prompted in English rather than in other languages \citep{muennighoff-etal-2023-crosslingual, ozsoy2024multilingualpromptsllmbasedrecommenders,koto-etal-2024-arabicmmlu}.

\subsection{Results and Analysis}
Table \ref{tab:models_performance} provides a comprehensive overview of model performance across all four tasks, consistently highlighting GPT-4o as the top-performing model. Among open-weight models, Gemma-2-9b-it outperforms others in the reordering task, while Qwen2.5-14B-Instruct achieves the highest accuracy in MCQ tasks, and Qwen2.5-7B demonstrates the strongest performance in the CB-QA task.  

Among models of comparable size, Qwen2.5 demonstrates superior performance relative to its counterparts. Notably, Mamba exhibits significantly weaker performance in procedural text comprehension compared to transformer-based models. Our findings indicate that increasing the number of parameters within the same model family generally enhances performance. Additionally, for language models with an available Instruct variant, the Instruct versions consistently achieve higher performance—except in the cases of Gemma-2-2B and Qwen2.5-1.5B.

We calculate Kendall rank correlation scores across four tasks to assess task sensitivity when comparing LLMs. Our analysis shows strong correlations (0.8--0.9) between reordering, PB-MCQ, and CB-MCQ, indicating that these tasks rank models similarly. However, CB-QA (the generation task) has a lower correlation (0.4--0.5) with the others, suggesting that text generation captures different aspects of procedural reasoning and is a valuable addition to the evaluation.

\begin{table*}[ht!]
  \centering
  \resizebox{0.8\textwidth}{!}{
  \begin{tabular}{p{4.7cm} wc{0.8cm} wc{0.8cm} wc{0.8cm} wc{1.5cm} wc{1.5cm} wc{0.8cm} wc{0.8cm} wc{0.8cm}}
    \toprule
    \multirow{2}{*}{\textbf{Model}} & \multicolumn{3}{c}{\textbf{Reordering}} & \multirow{2}{*}{\textbf{PB-MCQ}} & \multirow{2}{*}{\textbf{CB-MCQ}} & \multicolumn{3}{c}{\textbf{CB-QA}} \\
    \cmidrule{2-4} \cmidrule{7-9}
    & $\bm{\rho}\uparrow$ & \textbf{LD}$\downarrow$ & $\bm{\tau}\uparrow$ & & &  \textbf{R-L} &  \textbf{BS} &  \textbf{SS}\\
    \midrule
    Random & 0.00 & 5.56 & 0.00 & 0.25 & 0.25 & 0.00 & 0.00 & 0.00  \\
    \hdashline
    DeepSeek-R1\small(Distill-Llama-8B) & 0.30 & 5.48 & 0.14 & \centering 0.27 & \centering 0.38 & 0.20 & 0.53 & 0.48\\
    DeepSeek-R1\small(Distill-Qwen-14B) & 0.43 & 4.47 & 0.33 & \centering 0.42 & \centering 0.54 & 0.22 & 0.55 & 0.50 \\
    \hdashline
    Gemma-2-2b & 0.30 & 5.49 & 0.14 & \centering 0.27 & \centering 0.30  & 0.23 & 0.58 & 0.55 \\
    Gemma-2-2b-it & 0.21 & 5.48 & 0.10 & \centering 0.08 & \centering 0.26  & \textbf{0.25} & 0.55 & 0.48 \\
    Gemma-2-9b & 0.48 & 4.94 & 0.32 & \centering 0.37 & \centering 0.57  & 0.21 & 0.53 & 0.47 \\
    Gemma-2-9b-it & \textbf{0.75} & \textbf{3.09} & \textbf{0.66} & \centering 0.43 & \centering 0.46  & 0.15 & 0.50 & 0.42 \\
    \hdashline
    Llama-3.1-8B & 0.30 & 5.49 & 0.14 & \centering 0.28 & \centering 0.32  & 0.24 & \textbf{0.59} & 0.55  \\
    Llama-3.1-8B-Instruct & 0.43 & 4.69 & 0.33 & \centering 0.37 & \centering 0.48  & 0.24 & \textbf{0.59} & 0.55 \\
    Llama-3.2-1B & 0.29 & 5.56 & 0.12 & \centering 0.25 & \centering 0.27  & 0.22 & 0.57 & 0.52 \\
    Llama-3.2-1B-Instruct & 0.31 & 5.44 & 0.15 & \centering 0.26 & \centering 0.34  & 0.23 & 0.58 & 0.54 \\
    Llama-3.2-3B & 0.29 & 5.56 & 0.12 & \centering 0.26  & \centering 0.28  & 0.23 & 0.58 & 0.54 \\
    Llama-3.2-3B-Instruct & 0.25 & 5.35 & 0.14 & \centering 0.33 & \centering 0.36  & 0.24 & \textbf{0.59} & 0.54  \\
    \hdashline
    Mamba-1.4b-hf & 0.30 & 5.59 & 0.14 & \centering 0.00 & \centering 0.00  & 0.13 & 0.39 & 0.25 \\
    Mamba-2.8b-hf & 0.25 & 5.51 & 0.14 & \centering 0.00 & \centering 0.00  & 0.12 & 0.39 & 0.26 \\
    \hdashline
    Mistral-7B-Instruct-v0.2 & 0.51 & 4.60 & 0.40 & \centering 0.34 & \centering 0.48  & 0.23 & 0.58 & 0.54 \\
    Mistral-7B-v0.3 & 0.19 & 5.45 & 0.11 & \centering 0.32 & \centering 0.31  & 0.24 & 0.58 & 0.54 \\
    Mistral-7B-Instruct-v0.3 & 0.38 & 4.75 & 0.30 & \centering 0.37 & \centering 0.41  & 0.24 & \textbf{0.59} & 0.55 \\
    Mistral-Nemo-Base-2407 & 0.33 & 5.34 & 0.18 & \centering 0.34 & \centering 0.50  & 0.20 & 0.53 & 0.43 \\
    Mistral-Nemo-Instruct-2407 & 0.43 & 4.59 & 0.34 & \centering 0.39 & \centering 0.57  & 0.21 & 0.53 & 0.43\\
    \hdashline
    Qwen2.5-1.5B & 0.38 & 5.15 & 0.26 & \centering 0.31 & \centering 0.34  & 0.23 & 0.58 & 0.55 \\
    Qwen2.5-1.5B-Instruct & 0.42 & 4.82 & 0.32 & \centering 0.33 & \centering 0.21  & 0.23 & 0.58 & 0.54 \\
    Qwen2.5-7B & 0.63 & 3.94 & 0.52 & \centering 0.45 & \centering 0.54  & \textbf{0.25} & \textbf{0.59} & \textbf{0.57} \\
    Qwen2.5-7B-Instruct & 0.69 & 3.65 & 0.60 & \centering 0.50 & \centering 0.60  & 0.24 & \textbf{0.59} & 0.56 \\
    Qwen2.5-14B & 0.70 & 3.33 & 0.62 & \centering 0.48 & \centering 0.64  & 0.26 & \textbf{0.59} & 0.55 \\
    Qwen2.5-14B-Instruct & 0.72 & 3.21 & 0.64 & \centering \textbf{0.56} & \centering \textbf{0.70}  & 0.24 & \textbf{0.59} & 0.56 \\
    \hdashline
    Aya-Expanse-8b & 0.47 & 4.71 & 0.37 & \centering 0.39 & \centering0.43  & 0.24 & 0.58 & 0.54 \\
    Bloomz-560m & 0.37 & 5.39 & 0.22 & \centering 0.15 & \centering 0.00  & 0.14 & 0.45 & 0.32 \\
    Bloomz-7b1 & 0.29 & 5.48 & 0.13 & \centering 0.09 & \centering 0.20  & 0.17 & 0.50 & 0.38 \\
    mT0-xxl & 0.29 & 5.54 & 0.13 & \centering 0.35 & \centering 0.27  & 0.14 & 0.50 & 0.43\\
    \midrule
    GPT-4o & \textbf{0.81} & \textbf{2.38} & \textbf{0.75} & \centering 0.58 & \centering \textbf{0.74}  & \textbf{0.29} & \textbf{0.62} & \textbf{0.60}\\
    O3-mini & 0.78 & 2.54 & 0.72 & \centering\textbf{0.66} & \centering0.65 & 0.27 & 0.61 & 0.59\\
    \bottomrule
  \end{tabular}
  \vspace{-0.5cm}
  }
  \caption{Models' performance across tasks. Metrics include Spearman's Rank Correlation ($\bm{\rho}$) [-1,1], Levenshtein Distance (LD) [0,$\infty$], and Kendall's Tau Rank Correlation ($\bm{\tau}$) [-1,1] for Reordering; accuracy [0,1] for PB-MCQ and CB-MCQ; and ROUGE-F1 [0,1], BERT-F1 [0,1], and Semantic Similarity (SS) [0,1] for CB-QA. Higher values indicate better performance for all metrics except LD, where lower is better.}
  \label{tab:models_performance}
\end{table*}

\paragraph{Analysis of PB-MCQ Subtypes} Table \ref{tab:mcq_type_performance_shortlist} showcases the performance of the top models across PB-MCQ task for four distinct question types: The results reveal that antecedent affirmative (AA) questions are the easiest for language models, while subsequent affirmative (SA) questions are the most challenging. This suggests that models find it easier to reason about preceding steps than following ones. Interestingly, while prior studies \citep{truong-etal-2023-language,she-etal-2023-scone,kassner-schutze-2020-negated} indicate that negation generally weakens model performance in NLP tasks, this pattern does not hold in the procedural context. Subsequent negative (SN) leads to better performance, whereas affirmative negative (AN) questions result in a notable decline in accuracy.

\begin{table}[t]
  \centering
  \resizebox{0.9\linewidth}{!}{
  \begin{tabular}{p{4.4cm} wc{0.7cm} wc{0.7cm} wc{0.7cm} wc{0.7cm}}
    \toprule
    \multirow{2}{*}{\textbf{Model}} & \multicolumn{4}{c}{\textbf{PB-MCQ}} \\
    \cline{2-5}
    & \small \textbf{SA} &  \small \textbf{SN} &  \small \textbf{AA} &  \small \textbf{AN} \\
    \hline
    Gemma-2-9b-it & 0.31 & 0.51 & 0.56 & 0.36 \\
    Llama-3.1-8B-Instruct & 0.24 & 0.38 & 0.55 & 0.33 \\
    Mistral-Nemo-Instruct-2407 & 0.29 & 0.34 & 0.60 & 0.34 \\
    Qwen2.5-14B-Instruct & 0.45 & 0.53 & 0.69 & 0.55 \\
    Aya-Expanse-8b & 0.33 & 0.38 & 0.61 & 0.24\\
    GPT-4o & 0.46 & 0.57 & \textbf{0.73} & 0.54 \\
    O3-mini & \textbf{0.51} & \textbf{0.70} & 0.72 & \textbf{0.72} \\
    \bottomrule
  \end{tabular}
  }
  \caption{Model performance on PB-MCQ across question types. ``SA'', ``SN'', ``AA'', and ``AN'' denote Subsequent Affirmative, Subsequent Negative, Antecedent Affirmative, and Antecedent Negative, respectively.}
  \label{tab:mcq_type_performance_shortlist}
\end{table}

\begin{figure}[t]
  \includegraphics[width=0.9\columnwidth]{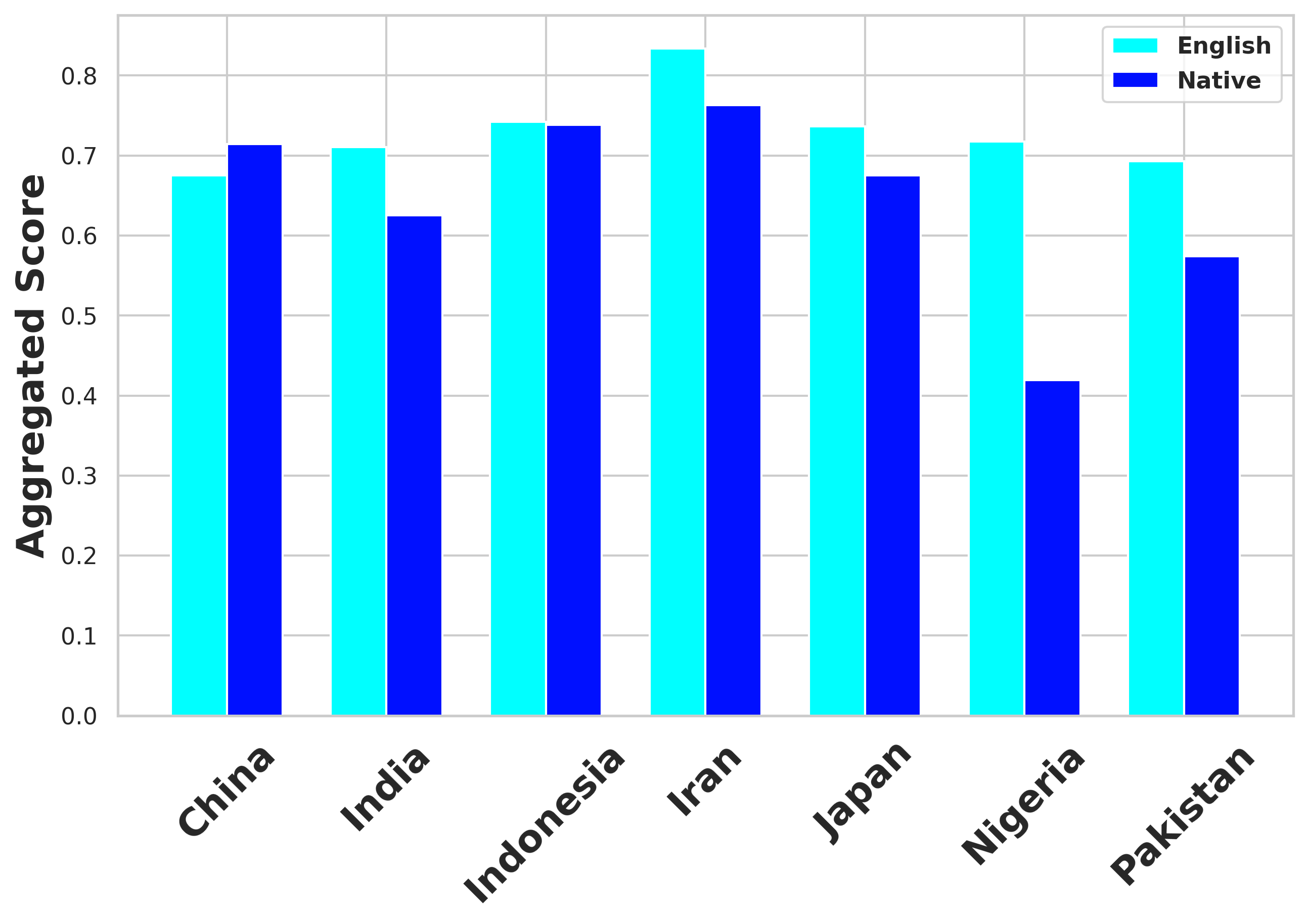}
  \caption{Language impact on Qwen2.5-14B-Instruct performance}
  \label{fig:language_impact_qwen}
  \vspace{-0.5cm}
\end{figure}

\noindent \textbf{Language Effects on Performance} We examine the influence of language on the performance of Qwen2.5-14B-Instruct, the top-performing open-weight model, in the reordering and PB-MCQ tasks. To quantify this effect, we first normalize the evaluation metrics for these tasks and then compute an aggregated score using a weighted sum approach.\footnote{The weighted sum assigns a weight of 0.6 to the reordering task (0.2 for each metric) and 0.4 to PB-MCQ accuracy.} As depicted in Figure \ref{fig:language_impact_qwen}, with the exception of China, Qwen2.5-14B-Instruct generally outperforms in English across other countries, especially in low-resource languages like Hausa and Urdu. This divergence may stem from the model’s extensive proficiency in English, whereas the linguistic nuances, idiomatic expressions, and procedural reasoning structures inherent to Chinese contexts might be underrepresented in the training data.

\begin{figure}[t]
  \includegraphics[width=\columnwidth]{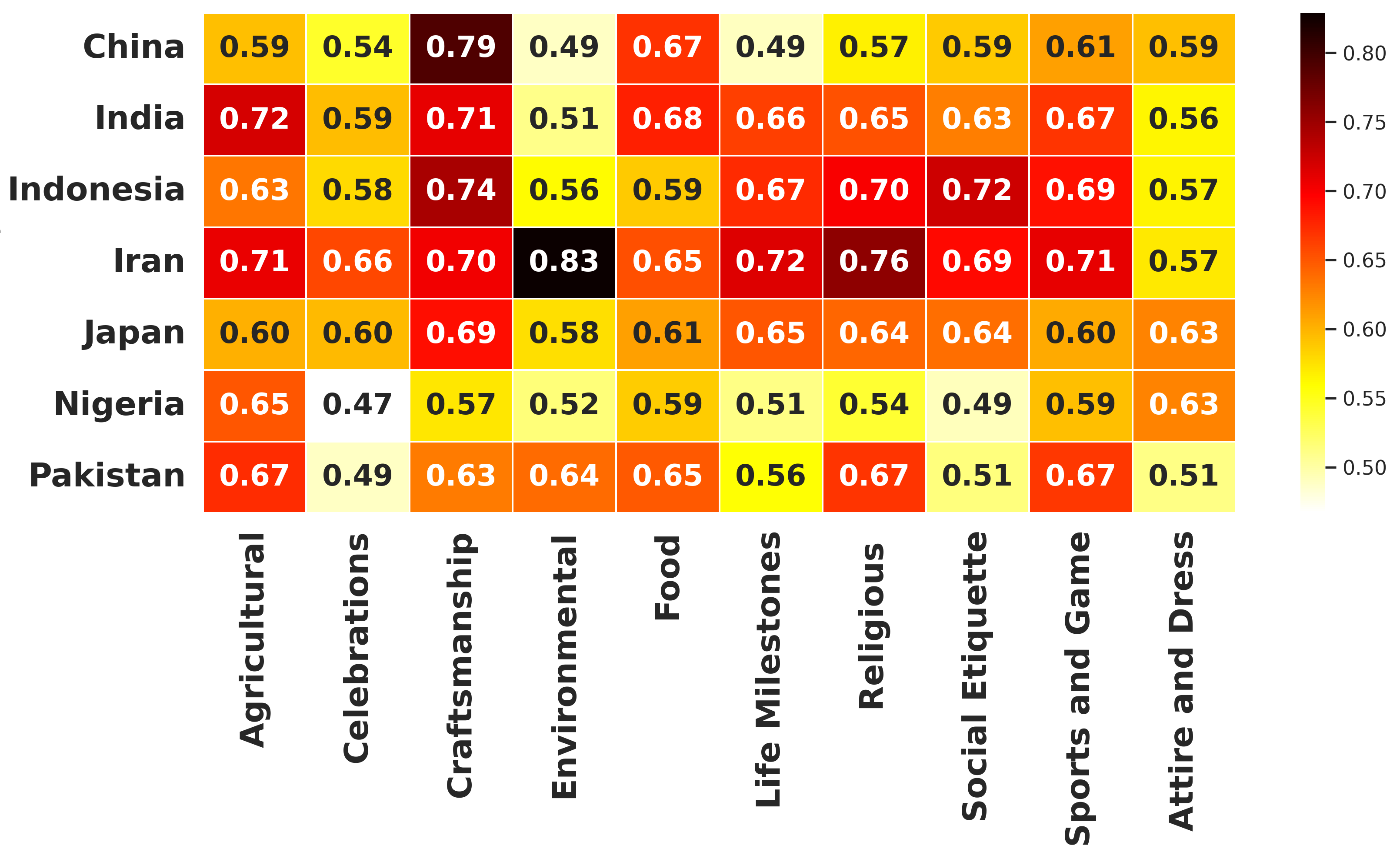}
  \caption{Cultural dimension performance by country}
  \label{fig:heatmap_country_category_qwen}
\end{figure}

\paragraph{Performance Across Cultural Dimensions} Figure~\ref{fig:heatmap_country_category_qwen} presents the aggregated performance scores of the top-performing open-weight model, Qwen2.5-14B-Instruct, across all evaluation tasks.\footnote{Scores are normalized and computed using a weighted sum approach, with the following weight distribution: 0.3 for reordering, 0.1 for each metric, 0.2 for PB-MCQ, 0.2 for CB-MCQ, and 0.3 for CB-QA.} Our analysis shows that Qwen2.5-14B-Instruct's knowledge of procedural texts varies across cultural domains. For instance, it demonstrates strong familiarity with Indian agriculture but performs less effectively on Chinese agricultural topics. Conversely, for craftsmanship and artisan skills, Qwen2.5-14B-Instruct encodes Chinese cultural knowledge better than other countries. In the food category, Indian cuisine is better represented, while Iranian religious practices appear more prominently. Additionally, Indonesian social etiquette are well captured by Qwen2.5-14B-Instruct, suggesting variation in how different cultural aspects are reflected in the model.

\begin{figure}[t]
  \includegraphics[width=\columnwidth]{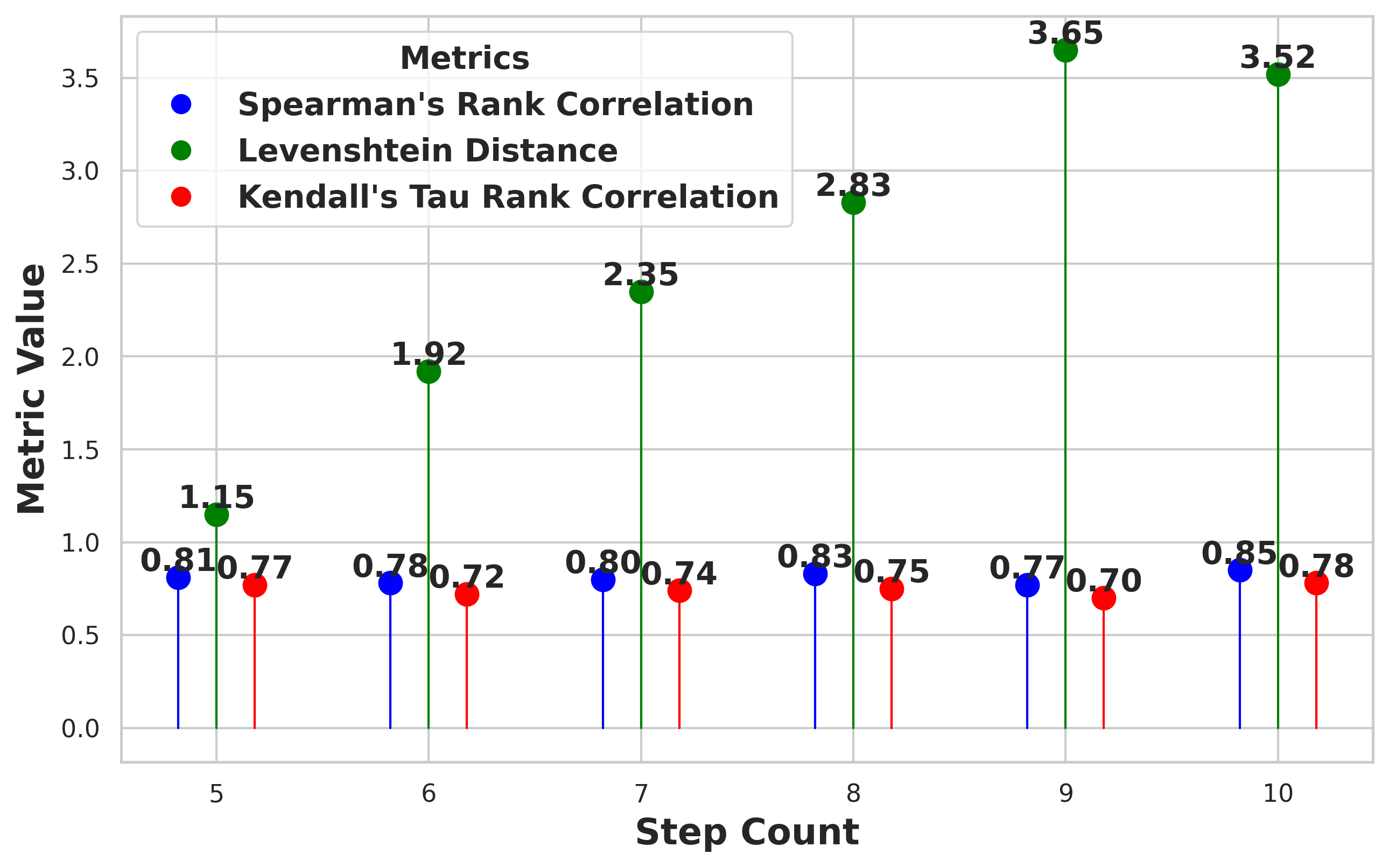}
  \caption{Impact of step count on reordering}
  \label{fig:step_count_impact}
\end{figure}

\paragraph{Impact of Procedure Length on Ordering} 
Figure \ref{fig:step_count_impact} shows that as procedure length increases, Levenshtein distance rises, indicating greater reordering difficulty. However, Spearman’s rank and Kendall’s Tau correlations remain high, suggesting that models generally preserve step order despite complexity. Shorter procedures introduce more ambiguity, making errors more impactful, while longer sequences benefit from stronger local dependencies, aiding order retention. Notably, procedures with ten steps achieve the highest rank correlations, indicating that structural cues in longer sequences enhance model performance. Most errors involve minor swaps rather than complete misordering.

\section{Conclusion}

We introduce \textbf{\texttt{CAPTex}}, a benchmark for evaluating mLLMs' ability to process culturally diverse procedural texts across seven languages. Our findings show that model performance varies across cultural domains, with greater challenges in tasks requiring implicit cultural knowledge, such as environmental practices, while structured domains like craftsmanship are better handled. Multiple-choice tasks in conversational contexts improve reasoning, while generation evaluation highlights gaps in procedural text comprehension.

\section*{Limitations}
Our study provides valuable insights into the performance of mLLMs on procedural texts, but there are a few limitations to consider. Firstly, our research is limited to textual data and does not include multimodal inputs, such as procedural texts with images. Incorporating images would enhance model understanding, but due to the added complexity, this is reserved for future work.

Additionally, our dataset focuses on seven countries, primarily due to budget constraints. While this may seem limited, the selected countries offer diverse language categories and varying resource levels, ensuring a meaningful analysis of cultural gaps in mLLMs. These findings can be generalized to a broader context, given the representativeness of the samples.

Finally, the conversation dataset consists of exchanges with exactly four utterances. While real-world dialogues are typically longer and more dynamic, this limitation was made for practical reasons. Despite the brevity, we found that mLLMs still struggle with maintaining coherence and understanding conversational flow, underscoring the challenges these models face, even in simpler dialogue settings.



\bibliography{acl_latex}

\appendix

\section{Dataset}
\label{sec:appendix}

\subsection{Detailed Categories}
\label{sec:appendix:categories}
\begin{enumerate} 
\item  \textbf{Food and Cuisine}: Captures cultural practices related to the preparation and consumption of food.
\item  \textbf{Celebrations and Festivals}: Represents rituals and activities associated with cultural events and festivities.
\item  \textbf{Social Etiquette and Hospitality}: Reflects societal norms and traditions in interpersonal interactions.
\item  \textbf{Craftsmanship and Artisan Skills}: Showcases traditional methods of creating culturally significant artifacts.
\item \textbf{Traditional Attire and Dress}: Highlights practices involving culturally significant clothing and adornments.
\item \textbf{Agricultural and Seasonal Practices}: Documents cultural procedures tied to agricultural cycles and seasonal changes.
\item \textbf{Religious and Spiritual Practices}: Encapsulates practices integral to religious and spiritual traditions.
\item \textbf{Life Milestones and Family Rites}: Represents cultural customs marking significant life events.
\item \textbf{Sports, Games, and Competitions}: Covers traditional recreational and competitive activities.
\item \textbf{Environmental and Nature-Based Practices}: Focuses on cultural interactions with and stewardship of the natural environment.
\end{enumerate}

\subsection{Procedures Collection}
\label{sec:appendix:data_collection}

In the initial phase, we conducted a pilot study using Prolific \footnote{\url{http://prolific.com/}} to recruit workers and gather some procedures. However, upon evaluating the quality of the collected procedures, we determined that Prolific was unsuitable for this specific task. Consequently, we opted to directly engage reliable workers who met our stringent requirements. Native speakers of the target languages were responsible for crafting each procedure in their native language. These were then precisely translated into English, ensuring both versions accurately conveyed the same content. Each procedure consisted of five to ten sequential steps, where the order was crucial for the proper understanding and execution of the tasks. To preserve the dataset's authenticity, workers were strictly prohibited from using AI-based text generation tools, which could introduce inaccuracies or fabrications, thereby compromising the reliability of the procedures. We implemented a two-step quality control process, comprising automated checks and peer evaluations through cross-verification. An all-inclusive checklist was employed to assess conceptual precision, cultural authenticity, logical flow, the necessity of maintaining step sequence, grammatical accuracy, and adherence to the required number of steps. This rigorous process ensured that both the native language and English versions met our high standards. 



\subsection{MCQ Design and Structure}
\label{sec:appendix:mcq_design}
As previously outlined, four distinct categories of questions are systematically generated using Python code. These questions adhere to the following templates for inquiries in English:
\begin{itemize}
\item Subsequent Affirmative: \textcolor{darkgray}{\textit{In the procedure “{Procedure Name}”, what is the next step after: “{Reference Step}”? }}
\item Subsequent Negative: \textcolor{darkgray}{\textit{In the procedure “{Procedure Name}”, which one is not the step before: “{Reference Step}”?}}
\item Antecedent Affirmative: \textcolor{darkgray}{\textit{In the procedure “{Procedure Name}”, which step must be completed before: “{Reference Step}”?}}
\item Antecedent Negative: \textcolor{darkgray}{\textit{In the procedure “{Procedure Name}”, which step does not come after: “{Reference Step}”?}}
\end{itemize}
These templates ensure clarity and consistency in question generation, aligning with the procedural framework.

\subsection{Conversation Generation}
\label{sec:appendix:conversation}
The following prompt is used to generate conversations using GPT-4o:
\textcolor{darkgray}{\textit{Create a short conversation between two people, Person A and Person B, based on the following procedure. The conversation should begin with (Have you heard about the “{Procedure Name}” from “{Country}”?) and progress naturally, with each message reflecting a clear and logical flow of ideas related to the steps of the procedure. The conversation should consist of four messages. In the third message, Person A asks a question about the next step in the procedure. In the last message, Person B should respond according to the procedure's step, providing a clear answer or action related to the next step, and explicitly mention the step number.\\\\
Example Structure:\\
Person A: Have you heard about the “{Procedure Name}” from “{Country}”?\\
Person B: (Response introducing the procedure and discussing some of the first steps)\\
Person A: (Asks a follow-up question to clarify the next step)\\
Person B: (Explains the first next step in the procedure according to the procedure’s step)\\
Next step: (number of next step)\\\\
IMPORTANT: Ensure the generated conversation adheres to the above structure. The last message should always be "Next step: (number)", where (number) is the next step in the procedure.\\\\
The procedure is as follows:\\
“{Procedure Steps}”
}}

Given the suboptimal performance of GPT-4o in processing lower-resource languages, this section is exclusively dedicated to generating conversations in English. 

The generated conversations are subjected to a two-phase verification process. The initial phase involves automated verification using a Python script, ensuring that each conversation comprises exactly four utterances. The subsequent phase entails thorough evaluation by qualified human annotators. These annotators assess each dialogue against a detailed checklist, which evaluates conceptual accuracy, grammatical precision, and alignment of Person B's responses with the procedural steps. During the review process, two types of errors were identified: 76 conversations had an incorrect next step number, and 45 conversations included explanations of multiple steps in the final utterance.

\section{Additional Results}
\label{sec:appendix:additional_results}
\begin{figure}[t]
  \includegraphics[width=\columnwidth]{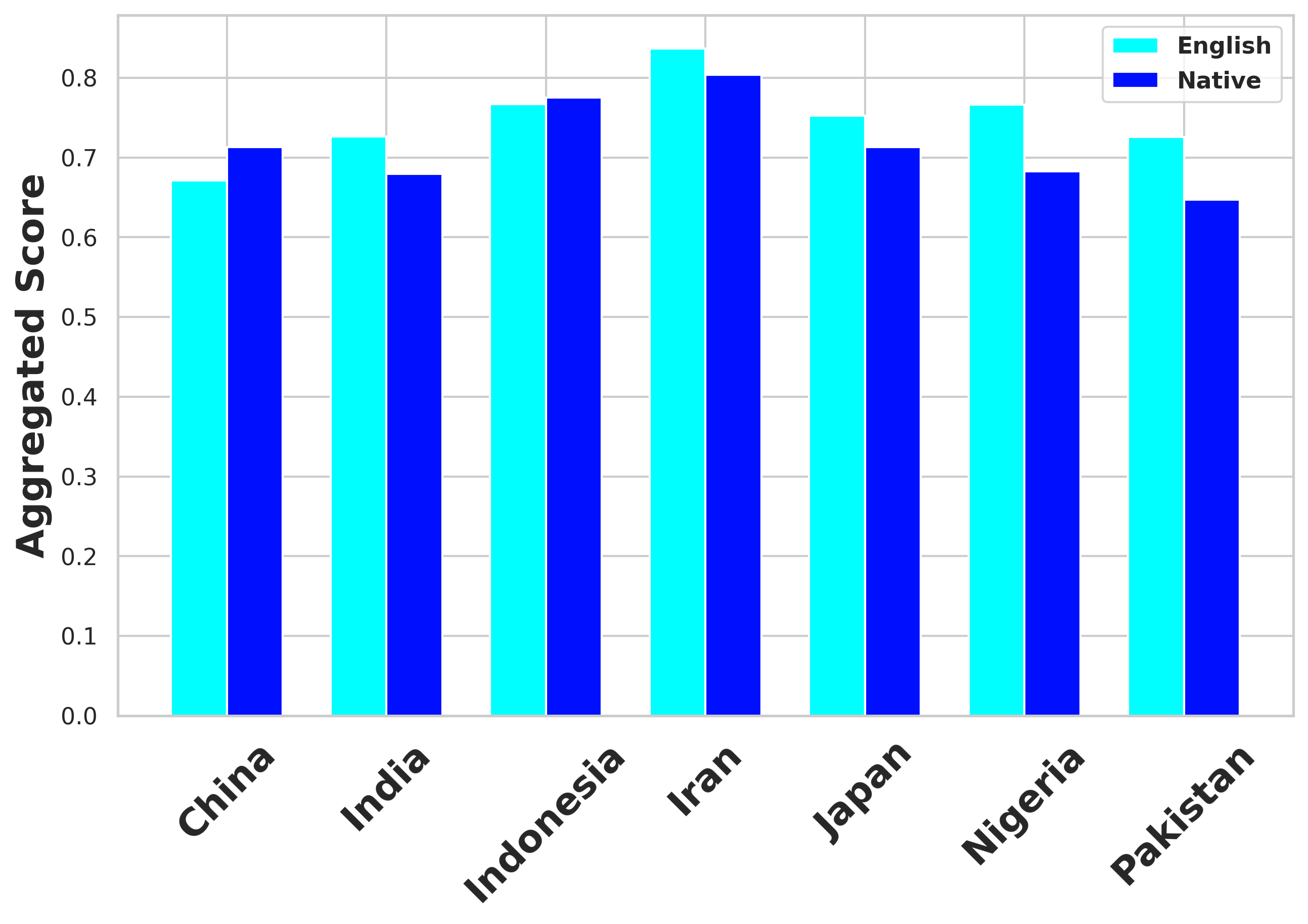}
  \caption{Language impact on GPT-4o performance}
  \label{fig:language_impact}
\end{figure}

\begin{figure}[t]
  \includegraphics[width=\columnwidth]{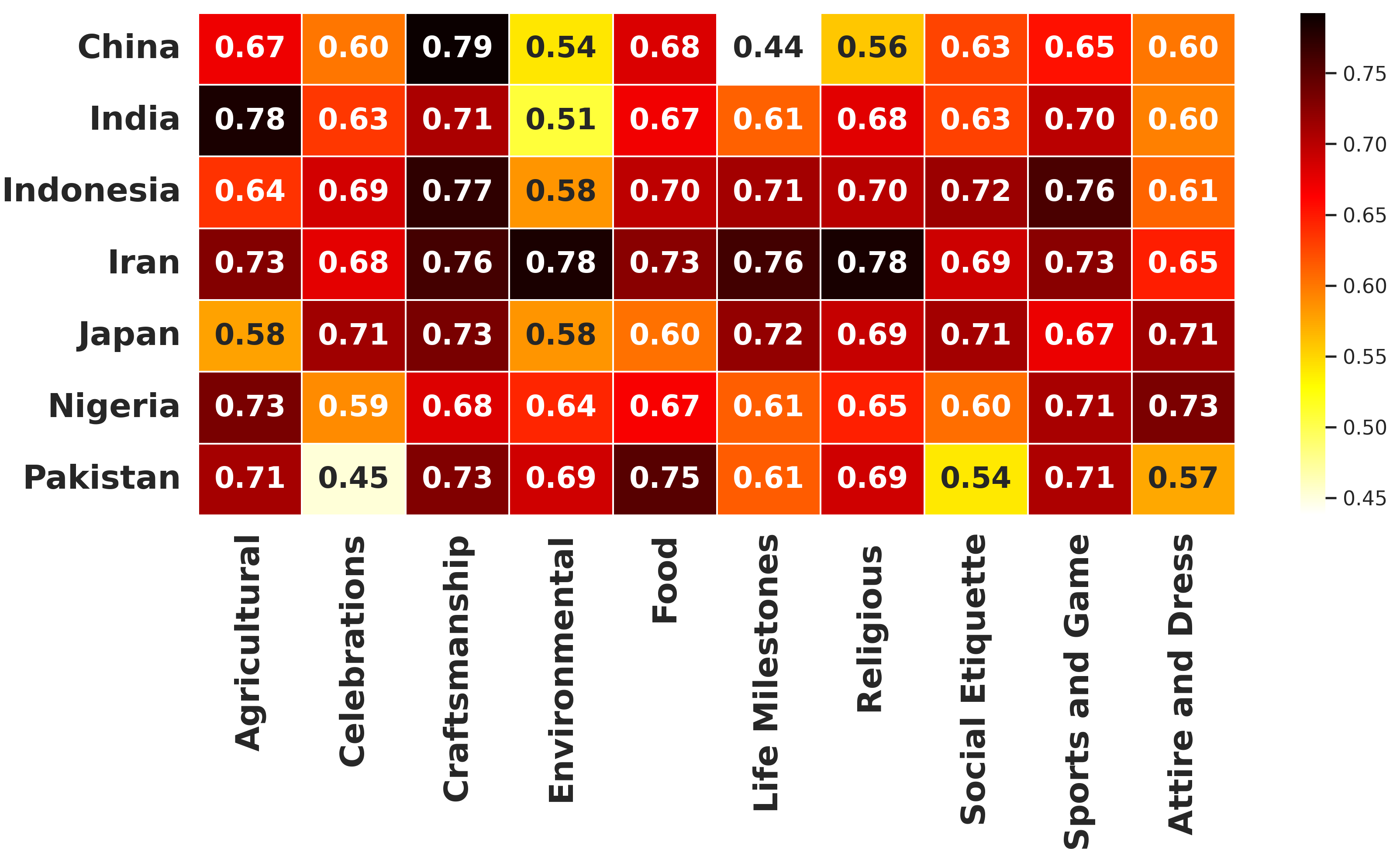}
  \caption{Cultural dimension performance by country}
  \label{fig:heatmap_country_category}
\end{figure}

\begin{figure*}[t]
\centering
  \includegraphics[width=2\columnwidth]{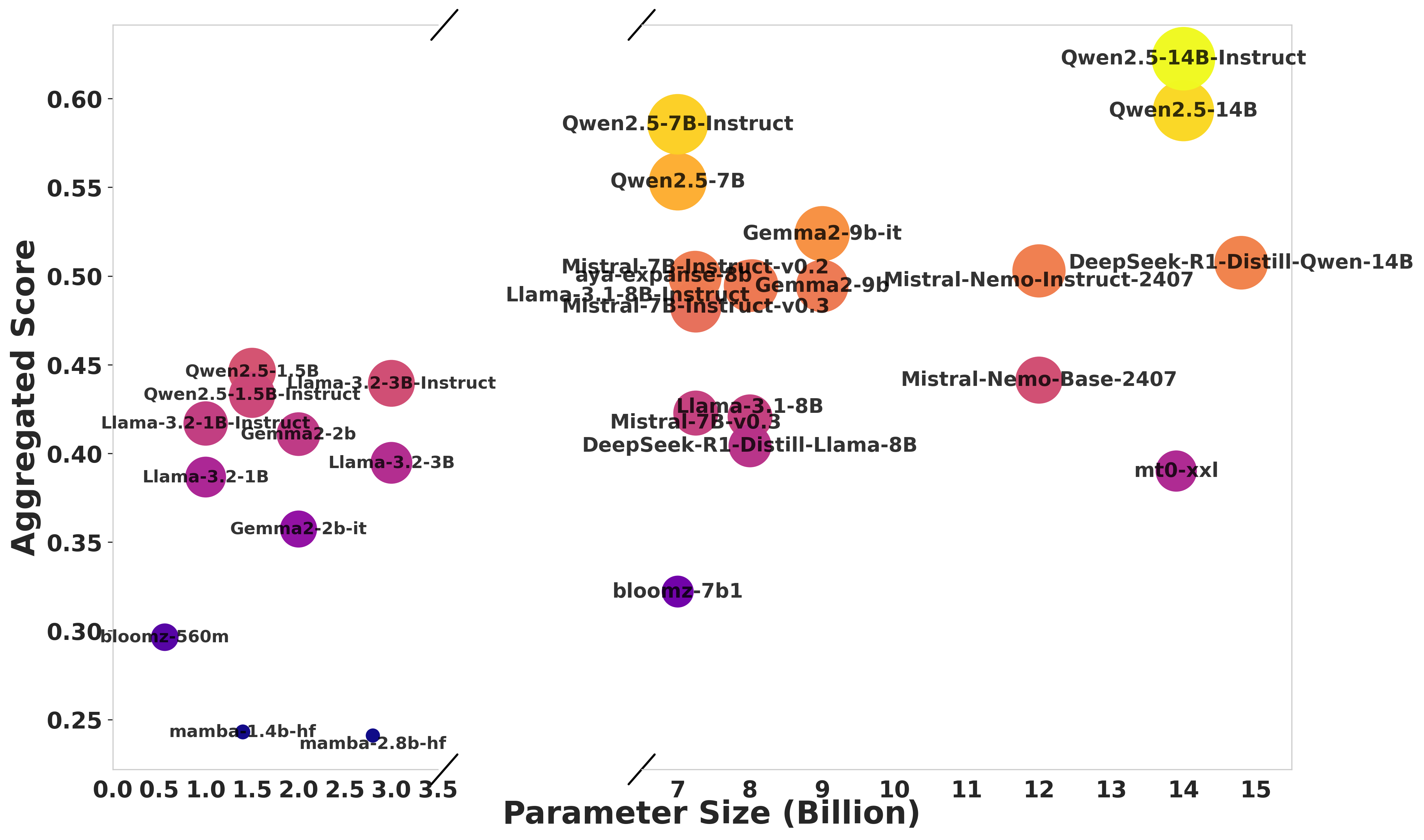}
  \caption{Performance of LLMs by model size}
  \label{fig:model_performance_bubble}
\end{figure*}

Figure \ref{fig:model_performance_bubble} presents the aggregated performance\footnote{Scores are normalized and computed using a weighted sum approach, with the following weight distribution: 0.3 for reordering, 0.1 for each metric, 0.2 for PB-MCQ, 0.2 for CB-MCQ, and 0.3 for CB-QA.} of the models across all four tasks. 
\paragraph{Language Effects on Performance} As shown in Figure \ref{fig:language_impact}, the results for GPT-4o, the top-performing model, are presented in a manner similar to those for Qwen2.5-14B-Instruct in Figure \ref{fig:language_impact_qwen}. GPT-4o generally performs better in English across most countries, with the exceptions of China and Indonesia. A comparison of the two figures reveals that GPT-4o maintains a more balanced performance between English and native languages in most countries. While GPT-4o tends to perform better in English, reflecting its extensive training on high-resource languages, it displays a smaller performance gap in low-resource languages such as Hausa (Nigeria) and Urdu (Pakistan). In contrast, Qwen2.5-14B-Instruct shows a more pronounced decline in native languages, indicating that GPT-4o may possess superior cross-lingual capabilities and stronger support for languages with limited resources. This gives GPT-4o a distinct advantage in multilingual contexts, while Qwen2.5-14B-Instruct shows a stronger preference for English, particularly in regions with fewer linguistic resources.
\paragraph{Performance Across Cultural Dimensions} The aggregated performance scores of GPT-4o, the top-performing model, are displayed in Figure \ref{fig:heatmap_country_category} for all evaluation tasks. Our analysis reveals that GPT-4o's understanding of procedural texts differs across cultural contexts. Specifically, the model demonstrates robust familiarity with Indian agricultural practices, while its performance on Japanese agricultural topics is comparatively weaker. In contrast, GPT-4o excels in encoding Japanese cultural knowledge related to celebrations and festivals, surpassing its representations of other cultures. In the culinary domain, Pakistani cuisine is more accurately captured, whereas Iranian religious practices are more prominently reflected. Furthermore, the model effectively encapsulates Indonesian social etiquette, highlighting the diversity in how various cultural elements are represented within the model.

\begin{table*}
  \centering
  \begin{tabular}{p{3.1cm} p{2cm} p{1.6cm} p{1.6cm} p{1.6cm} p{0.01cm} p{0.4cm} p{0.6cm} p{0.5cm} p{0.6cm}}
    \toprule
     \multirow{2}{*}{\textbf{Category}} &  \multicolumn{1}{c}{\textbf{Procedures}} & \multicolumn{3}{c}{\textbf{MCQs} \footnotesize (English/Native)} & \multicolumn{5}{c}{\quad\textbf{Conversations}\small(Utternaces)} \\
    \cline{3-5} \cline{7-10}
    & \footnotesize (English/Native) & \small \textbf{Question} & \small \textbf{Correct \newline Answer} & \small \textbf{Wrong \newline Answer} & & \small \textbf{First} & \small \textbf{Second} & \small \textbf{Third} & \small \textbf{Fourth} \\
    \hline
    Agricultural and Seasonal Procedures     & 94.6 / 104.1 & 26.8 / 32.7 & 13.6 / 14.6 & 13.4 / 14.3 & & 10.4 & 41.1 & 12.7 & 28.5  \\
    Celebrations and Festivals     & 118.2 / 124.2   & 31.7 / 35.8 & 17.4 / 17.9   & 17.8 / 18.2 & & 11.0 & 43.3 & 12.5 & 29.0 \\
    Craftsmanship and Artisan Skills & 132.4 / 146.5   & 32.6 / 39.0 & 18.6 / 20.5 & 18.7 / 20.3 & & 10.3 & 45.0 & 13.2 & 30.6 \\
    Environmental and Nature-Based Procedures      & 93.3 / 108.8 & 29.1 / 36.0 & 14.2 / 16.3 & 15.3 / 17.4 & & 11.3 & 42.5 & 12.9 & 28.8 \\
    Food and Cuisine     & 137.8 / 139.9 & 32.2 / 36.3 & 18.2 / 18.6 & 18.5 / 18.5 & & 10.7 & 43.2 & 13.1 & 28.6 \\
    Life Milestones and Family Rites   & 132.3 / 137.0 & 32.8 / 38.7 & 18.0 / 18.0 & 18.7 / 19.3 & & 11.7 & 43.3 & 12.7 & 29.5 \\
    Religious and Spiritual Practices  & 105.1 / 109.7 & 30.4 / 34.4 & 15.2 / 15.3  & 15.7 / 16.0 & & 12.1 & 41.4 & 12.5 & 28.0 \\
    Social Etiquette and Hospitality  & 115.0 / 125.2 & 32.8 / 38.7 & 17.2 / 18.6  & 18.5 / 19.7 & & 13.0 & 43.5 & 13.3 & 29.6 \\
    Sports, Games, and Competitions  & 112.2 / 122.1 & 30.5 / 36.0 & 16.1 / 16.9  & 16.9 / 18.3 & & 10.5 & 42.6 & 13.6 & 28.2 \\
    Traditional Attire and Dress  & 132.5 / 131.8 & 32.7 / 36.2 & 18.0 / 17.9  & 18.4 / 18.1 & & 11.6 & 43.6 & 12.9 & 31.1 \\
    \bottomrule
  \end{tabular}
  \caption{Average word counts for \textbf{\texttt{CAPTex}} components (Procedures, MCQs, and Conversations) by category}
  \label{tab:word_count_category}
\end{table*}

\begin{CJK}{UTF8}{min} 
\begin{table*}
  \centering
  \renewcommand{\arraystretch}{1.5}
  \begin{tabular}{p{1.3cm} p{2.7cm} p{1.6cm} p{0.6cm} p{3cm} p{5cm}}
    \toprule
    \textcolor{teal}{\textbf{Country}} & \textcolor{magenta}{\textbf{Category}} & \textcolor{cyan}{\textbf{Language}} & \textcolor{olive}{\textbf{Type}} & \textbf{Question} & \textbf{Choices} \\
    \hline
    Japan & Food and Cuisine & Japanese & \centering SA & 手順 '昆布巻きの作り方（4人前）' では、'鍋に6と5の戻し汁を入れ、中火で沸騰させる。' の後の次のステップは何ですか？ & {A.} 1を昆布と戻し汁に分ける。昆布を端からしっかり巻き付けて爪楊枝で留める。\newline
    {B.} 4を巻き付け、きつく結んだら爪楊枝を抜き、余分なかんぴょうを切る。\newline \textcolor{green}{C. 醤油大さじ1.5、顆粒和風だし大さじ1、料理酒大さじ1、みりん大さじ2、砂糖大さじ2、塩小さじ1/2を加える。}\newline {D.} 塩を洗い流し、水気を切ったら20cm幅に切り、12本にする。\\ \hline
    Japan & Food and Cuisine & English & \centering SA &  In the procedure 'How to Make Kelp Rolls (for 4 Persons)', what is the next step after: 'Place the rolls into a saucepan and add the stock. Bring to a boil over medium heat.'? & {A.} Separate the kelp and stock. Wrap the kelp tightly around the filling and secure it with a toothpick. \newline
    {B.} Wrap a piece of kanpyo around the roll and tie it tightly. Remove the toothpick and trim any excess kanpyo. \newline \textcolor{green}{C. Add 1.5 tbsp soy sauce, 1 tbsp granulated Japanese soup stock, 1 tbsp cooking sake, 2 tbsp mirin, 2 tbsp sugar, and 1/2 tsp salt.} \newline {D.} Rinse off the salt, drain, and cut the kanpyo into 20 cm lengths, then divide into 12 pieces.\\
    \bottomrule
  \end{tabular}
  \caption{\label{tab:mcq_example}
    Example of MCQs
  }
\end{table*}
\end{CJK}

\begin{table*}
  \centering
  \renewcommand{\arraystretch}{1.5}
  \begin{tabular}{p{1.5cm} p{3cm} p{10cm}}
    \toprule
    \textcolor{teal}{\textbf{Country}} & \textcolor{magenta}{\textbf{Category}} & \textbf{Conversation} \\
    \hline
    Iran & Sports, Games, and Competitions & Person A: Have you heard about the Alak-Dolak Game from Iran?\newline
    Person B: Yes, it's quite an interesting traditional game! It starts with forming two teams, and they draw lots to see which team gets to go first. Each team also has a leader who plays a crucial role at the beginning. For instance, the leader places two stones and then positions the "Alak" stick on top of them while holding the "Dolak" stick in their hand.\newline
    Person A: That's fascinating! What does the leader do next after placing the Alak stick on the stones?\newline
    Person B: After placing the Alak stick, the leader uses the Dolak stick by positioning its tip under the Alak stick. Then they move the Dolak stick left and right to launch the Alak stick into the air. \newline
    Next step: 3.\\
    \hline
    Nigeria & Celebrations and Festivals & Person A: Have you heard about the Osun-Osogbo Festival from Nigeria?  \newline
    Person B: Yes, I have! It's a fascinating cultural event that spans two weeks, starting with cleansing rituals called Iwopopo. The festival begins by preparing the Arugba, a virgin maiden who carries a sacred calabash. She plays a vital role as she represents purity and connects the people with the goddess Osun. It's all part of cleansing the community of evil spirits.   \newline
    Person A: That sounds amazing. What happens after the cleansing rituals?  \newline
    Person B: After the cleansing, traditional Yoruba music, drumming, and dances are performed daily, highlighting the community's rich cultural heritage. Craftsmen and vendors set up in the marketplace near the sacred Osun Grove to display arts, crafts, and local foods.  \newline
    Next step: 2\\
    \bottomrule
  \end{tabular}
  \caption{\label{tab:conversation_example}
    Examples of a procedurally grounded conversations
  }
\end{table*}

\end{document}